\RequirePackage{fix-cm}
\documentclass[smallextended]{svjour3}       % onecolumn (second format)
\smartqed  % flush right qed marks, e.g. at end of proof
\usepackage{graphicx}
\usepackage{amsfonts}
\usepackage{diagbox}
\usepackage{multirow}
\usepackage{booktabs}
\usepackage{url}
\usepackage{natbib}
\usepackage{color} % for revision
\usepackage{soul} % for revision
\newcommand{\argmin}{\mathop{\rm argmin}\limits}
\newcommand{\sign}{\mathop{\rm sign}\limits}

\newcommand{\disp}{\displaystyle}

\newcommand*{\affaddr}[1]{#1} % No op here. Customize it for different styles.
\newcommand*{\affmark}[1][*]{\textsuperscript{#1}}

\begin{document}

\title{Binary classification with ambiguous training data %\thanks{Grants or other notes
%about the article that should go on the front page should be
%placed here. General acknowledgments should be placed at the end of the article.}
}
%\subtitle{Do you have a subtitle?\\ If so, write it here}

%\titlerunning{Short form of title}        % if too long for running head

\author{Naoya Otani\affmark[1]        \and
        Yosuke Otsubo\affmark[1] \and Tetsuya Koike\affmark[1] \and Masashi Sugiyama\textsuperscript{2,3} %etc.
}

\authorrunning{ Naoya Otani \and Yosuke Otsubo \and Tetsuya Koike \and Masashi Sugiyama} % if too long for running head

\institute{Naoya Otani \at
              Nikon Corporation, Research \& Development Division\\
              471, Nagaodai-cho, Sakae-ku, Yokohama-city, Kanagawa, 244-8533, Japan.\\
              Tel.: +81-45-853-8584\\
              Fax: +81-45-853-8585\\
              \email{Naoya.Otani@nikon.com}          \\ \\
             \affaddr{\affmark[1] Nikon Corporation, Research \& Development Division}\\
             \affaddr{\affmark[2] RIKEN Center for Advanced Intelligence Project}\\
             \affaddr{\affmark[3] The University of Tokyo, Graduate School of Frontier Sciences } 
}

\date{Received: date / Accepted: date}
% The correct dates will be entered by the editor

\maketitle

\begin{abstract} In supervised learning, we often face with \emph{ambiguous} (A) samples that are difficult to label even by domain experts.
In this paper, we consider a binary classification problem in the presence of such A samples.
This problem is substantially different from semi-supervised learning since unlabeled samples are not necessarily difficult samples.
Also, it is different from 3-class classification with the positive (P), negative (N), and A classes since we do not want to classify test samples into the A class.
Our proposed method extends binary classification with reject option, which trains a classifier and a rejector simultaneously using P and N samples based on the 0-1-$c$ loss with rejection cost $c$.
More specifically, we propose to train a classifier and a rejector under the 0-1-$c$-$d$ loss using P, N, and A samples, where $d$ is the misclassification penalty for ambiguous samples.
In our practical implementation, we use a convex upper bound of the 0-1-$c$-$d$ loss for computational tractability.
Numerical experiments demonstrate that our method can successfully utilize the additional information brought by such A training data.

%教師あり学習は機械学習における代表的な手法であり、工学的な観点でも重要な技術である。しかしながら、実際に教師あり学習でモデルを学習するときに、大量のデータを一定の基準でラベル付けするのは、非常に困難である。そこで我々は、アノテータが判断に確信を持てないときに、与えられたラベル以外の「Ambiguous」なラベルを許すことを考えた。本論文で、我々は曖昧なラベルを考慮した新規の損失関数を提案し、これに対応する代理損失を導出した。更に、代理損失のハイパーパラメータに関して、理論的な最適値を導出した。最後に数値実験を行い、数理モデルおよび実データで本手法の有効性を示した。
% \PACS{PACS code1 \and PACS code2 \and more}
% \subclass{MSC code1 \and MSC code2 \and more}
\end{abstract}

\section{Introduction}
Supervised learning has been successfully deployed in various real-world applications, such as medical diagnosis \citep{bar2015chest,wang2016deep,esteva2017dermatologist} and manufacturing systems \citep{park2016machine,ren2017generic}.
However, when the amount of labeled data is limited, current supervised learning methods still do not work reliably \citep{pesapane2018artificial}.

To efficiently obtain labeled data, domain knowledge has been used in many application areas \citep{ren2017generic,cruciani2018automatic,konishi2019practical,bejnordi2017diagnostic}.
However, as some studies have pointed out \citep{wagner2005physiological,li2016facial,shahriyar2018approach}, there are often \emph{ambiguous} samples that are substantially difficult to label even by domain experts.

% 2020/06/25 revise
%When we face hard-to-label samples in labeling processes, we naively have two approaches:
%one is to skip labeling such samples, and another is to forcibly give either label.
%But, we point out that neither approach handles such samples appropriately, since the former can give biases to the distribution of the training data, and the latter can give inconsistent labels those samples.
%Therefore, we consider the third approach, which is to give the ambiguous label to such samples.
%However, to the best of our knowledge, no method has been proposed that can effectively utilize such ambiguous data for binary classification.
% 2020/06/25 end_revise
The goal of this paper is to propose a novel classification method that can handle such ambiguous data.
More specifically, we consider a binary classification problem where, in addition to positive (P) and negative (N) samples, ambiguous (A) samples are available for training a classifier.

Naively, we may consider employing 3-class classification methods for the P, N, and A classes.
However, since we classify test samples only in the P or N class, not in the A class, naive 3-class methods cannot be directly used in our problem. Moreover, they cannot utilize the information that the A class exists between the P and N classes.
Another related approach is classification with reject option \citep{bartlett2008classification,cortes2016learning}, where ambiguous test samples are not classified into the P or N classes, but rejected.
However, classification methods with reject option do not consider ambiguous samples in the training phase and thus they cannot be employed in the current scenario.

Semi-supervised learning may also be related to the current problem, where unlabeled data is used for training a classifier in addition to P and N data \citep{odena2016semi,sakai2017semi}.
In semi-supervised learning, unlabeled samples are P and N samples that have not yet been labeled and they are not necessarily difficult samples to be labeled.
On the other hand, in our target problem of classification with ambiguous data, ambiguous data are typically distributed in the intersection of the P and N classes.
Thus, since the problem setups are intrinsically different, merely using semi-supervised learning methods in the current problem may not be promising.

Classification with imperfect labeling \citep{cannings2020classification} allows incorrect labels in training data, so it can be useful to deal with a dataset where annotators forcibly give positive or negative labels to all samples.
Open-set classification \citep{scheirer2012toward} detects samples classified into none of the training classes and classifies them into the ``unknown'' class, so we can apply it when we deal with a dataset where annotators skip labeling hard-to-label samples and we classify test samples into the P, N or A class.
However, those two approaches are still different from our problem setting in terms of labels in input and output data.
Table~\ref{tab} summerizes the problem setting of different approaches.

\begin{table}
\centering
\caption{Problem settings of related and our methods.}
\begingroup
\renewcommand{\arraystretch}{1.1}
\scriptsize
\begin{tabular}{llll}
\hline \\[-8pt]
\begin{tabular}{l}Methods\end{tabular} & \begin{tabular}{l}Labels in\\training data\end{tabular} & \begin{tabular}{l}Labels predicted\\in test phase\end{tabular} & \begin{tabular}{l}Relationship\\among classes\end{tabular}\\
\hline
\hline
\begin{tabular}{l}Binary \\classification\end{tabular} & \begin{tabular}{ll}Positive\\Negative\end{tabular} & \begin{tabular}{ll}Positive\\Negative\end{tabular} & \begin{tabular}{l}None\end{tabular}\\
\hline
\begin{tabular}{l}3-class \\classification\end{tabular} & \begin{tabular}{ll}Class 1\\Class 2\\Class 3\end{tabular} & \begin{tabular}{ll}Class 1\\Class 2\\Class 3\end{tabular}&\begin{tabular}{l}None\end{tabular} \\
\hline
\begin{tabular}{l}Classification \\with reject \\option\end{tabular}& \begin{tabular}{ll}Positive\\Negative\end{tabular} & \begin{tabular}{ll}Positive\\Rejected\\Negative\end{tabular} &\begin{tabular}{l}Rejected samples \\are in P/N mixed\\ regions\end{tabular} \\
\hline
\begin{tabular}{l}Semi-supervised\\learning\end{tabular}& \begin{tabular}{ll}Positive\\Unlabeled\\Negative\end{tabular} & \begin{tabular}{ll}Positive\\Negative\end{tabular} &\begin{tabular}{l}Unlabeled samples \\belong to P or N\end{tabular}  \\
\hline
\begin{tabular}{l}Classification \\ with imperfect \\labeling\end{tabular} & \begin{tabular}{ll}Positive\\Negative\end{tabular} & \begin{tabular}{ll}Positive\\Negative\end{tabular} &\begin{tabular}{l}Training data can\\ contain incorrect \\labels\end{tabular} \\
\hline
\begin{tabular}{l}Open-set \\ classification\end{tabular} & \begin{tabular}{ll}Positive\\Negative\end{tabular} & \begin{tabular}{ll}Positive\\Negative\\Unknown\end{tabular} &\begin{tabular}{l}Test data can\\contain neigher\\positive nor\\negative samples\end{tabular} \\
\hline
\begin{tabular}{l}Our proposal\end{tabular} & \begin{tabular}{ll}Positive\\Ambiguous\\Negative\end{tabular} & \begin{tabular}{ll}Positive\\Negative\end{tabular} &\begin{tabular}{l}Ambiguous \\samples are in \\P/N mixed regions\end{tabular} \\
\hline
\end{tabular}\label{tab}
\endgroup
\end{table}

To effectively solve the problem of classification with ambiguous data, we propose to extend classification with reject option
that trains a classifier and a rejector simultaneously using P and N samples based on the 0-1-$c$ loss with rejection cost $c$ \citep{cortes2016learning}.
Our proposed method trains a classifier and a rejector under the 0-1-$c$-$d$ loss using P, N, and A samples, where $d$ is the misclassification penalty for ambiguous samples.
Then, in the test phase, we use the trained classifier to assign P or N labels to test samples.
However, in the same way as the 0-1-$c$ loss, directly performing optimization with the 0-1-$c$-$d$ loss is cumbersome due to its discrete nature.
To cope with this problem, we introduce a convex upper bound of the 0-1-$c$-$d$ loss and use it in our practical implementation.
Through experiments, we demonstrate that the proposed method can improve the test classification accuracy by utilizing A samples in the training phase.
We also consider a simple heuristic that we randomly relabel ambiguous samples into the positive or negative class and apply classification with reject option.
We show that the heuristic is essentially equivalent to a special case of the proposed method, and thus it can be an easy-to-implement alternative to the proposed method.% さらに、不明サンプルをランダムにP/Nに再ラベリングしてリジェクト付き分類を学習した方法が、提案手法と等価であることを示し、より実装しやすい代理の方法を提案した。

The rest of this paper is organized as follows.
%In Section 2, we review works that are related to classification with ambiguous data.
%Then we describe the details of our proposed method in Section 3 and experimentally evaluate its performance in Section 4.
We briefly review supervised learning in Section 2.
Then, we define our problem setting and describe the details of our proposed method in Section 3.
In Section 4, we experimentally evaluate its performance.
Finally, in Section 5, we summarize our contributions and describe future works.

\section{Supervised classification}
In this section, we first define the standard supervised classification problem and
then review its standard solution.

\subsection{Formulation}
Let $x \in \mathcal{X}$ be an input point and $y \in \mathcal{Y}=\{1,-1\}$ denote a binary label, which corresponds to the positive and negative classes, respectively. Suppose that we are given a set of positive and negative samples $\{(x_i, y_i)\}_{i=1}^N$ drawn independently from the probability distribution with density $p(x,y)$ defined on $\mathcal{X}\times\mathcal{Y}$.
Let $h: \mathcal{X}\rightarrow \mathbb{R}$ denote a discriminant function, with which a class label is predicted for test input point $x$ as $\hat{y} = \sign(h(x))$.

The goal is to train the discriminant function $h$ so that the expected misclassification rate is minimized. Let us define the 0-1 loss as
\begin{equation}
L_{01}(h,x,y) = 1_{yh(x)\leq0}, \label{0-1loss}
\end{equation}
where $1_A$ is the indicator function that takes 1 if statement $A$ is true and 0 otherwise.
Then, we can express this problem as
\begin{eqnarray*}
h^* &=& \argmin_{h} R(h), \\
R(h) &=& \mathbb{E}_{p(x,y)} \left[ L_{01}(h, x, y) \right] ,
\end{eqnarray*}
where $h^*$ denotes the optimal discriminant function
and $\mathbb{E}_{p(x,y)}$ denotes the expectation over $p(x,y)$.
In practice, since we do not know the true density $p(x,y)$, we usually use the empirical distribution to approximate the expectation:
\begin{equation}
\hat{R}(h) = \frac{1}{N} \sum_{i=1}^{N} L_{01}(h, x_i, y_i). \label{empiricalrisk}
\end{equation}

Based on Eqs.~(\ref{0-1loss}) and (\ref{empiricalrisk}), we can formulate various classification methods depending on loss functions \citep{book:Sugiyama:2015}.
In the rest of this section, we introduce the support vector machine (SVM) \citep{book:Vapnik:1995}, which is one of the most basic algorithms of binary classification.

\subsection{Support vector machine (SVM)}
Because optimization with $L_{01} (h,x,y)$ is computationally intractable, it is not practical to optimize the empirical risk $\hat{R}(h)$ directly. To overcome this problem,
the hinge loss, an upper bound of $L_{01}(h,x,y)$ called the hinge loss, defined by
\begin{equation}
L_\mathrm{H}(h,x,y)=\max\left(1-yh(x),0\right), \label{L_H}
\end{equation}
has been introduced as its surrogate. Since the hinge loss is convex, optimization can be reduced to a convex program. Further introducing the L2 regularization, basis functions $\phi_1(x),\ldots,\phi_N(x)$, and slack variables $\xi=(\xi_1,\ldots,\xi_N)^\top$ with $^\top$ being the transpose, the following quadratic program can be obtained as a dual optimization problem:
%Thus, we can minimize the corresponding empirical loss
%\begin{equation}
%\hat{R}_\mathrm{H}(h)=\frac{1}{N}\sum_{i=1}^{N}L_\mathrm{H}(h,x_i,y_i)
%\end{equation}
%by some gradient methods.
%
%We define the concrete model of the discriminant function. A positive-definite kernel $K$ is applied to represent the discriminant function $h$ like
%\begin{equation}\label{kernel}
%h(x;w)=\sum_{i=1}^{N}w_iK(x,x_i),
%\end{equation}
%then the optimal discriminant function is expressed as
%\begin{eqnarray}
%h^*(x) = h(x;w^*) \nonumber \\
%w^* = \argmin_w\left[ R_\mathrm{H}\left(h(x;w)\right)+\frac{\lambda}{2N}\sum_{j=1}^{N}w_j^2 \right] \label{minimization_w}
%\end{eqnarray}
%where we applied L2 regularization. Polynomial kernel
%\begin{equation}
%K_\mathrm{POL}(x_a, x_b; n) = (1+x_a\cdot x_b)^n
%\end{equation}
%or RBF kernel
%\begin{equation}
%K_\mathrm{RBF}(x_a, x_b; \sigma) = \frac{1}{\sqrt{2\pi\sigma^2}}\exp\left[ -\frac{\left\| x_a - x_b \right\|^2}{2\sigma^2}\right]
%\end{equation}
%are usually applied as the positive-definite kernel $K$.
%We introduce a slack variable $\xi$ as
%\begin{equation}
%L_\mathrm{H}(h,x,y) = \xi \qquad(\xi \geq 1-yh(x), \quad \xi\geq 0)
%\end{equation}
%so that we replace the $\max$ function of inequality constraints. Then we reduce Eq. (\ref{minimization_w}) to a quadratic programming problem
\begin{eqnarray}
(\hat{w}, \hat{\xi}) &=& \argmin_{(w, \xi)}\left[\frac{\lambda}{2} \| w \|^2 + \frac{1}{N} \sum_{i=1}^{N}\xi_i \right] \nonumber \\
&&\mathrm{s.t.} \left(
\begin{array}{c}
\xi_i \geq 1 - y_i h_i\\
\xi_i \geq 0
\end{array}
\right) \mathrm{for}~i=1,\ldots,N,
\end{eqnarray}
where $w=(w_1,\ldots,w_N)^\top$ are the coefficients of the discriminant function, $\lambda>0$ is the L2 regularization parameter,
and $h_i$ is the value of the discriminant function at sample point $x_i$ given by
$h_i=\sum_{j=1}^{N}w_j\phi_{j}(x_i)$.
The resulting discriminant function is given by $h(x; \hat{w}) = \sum_{j=1}^N \hat{w}_j \phi_j(x)$.

\section{Classification with ambiguous data}\label{sec:cad}
In this section, we formulate our target problem
called \emph{classification with ambiguous data} (CAD)
and propose a new method for solving the CAD.

\subsection{Formulation}
We consider three class labels, i.e,. positive, ambiguous, and negative: $y \in \mathcal{Y}_0=\{1,0,-1\}$.
Suppose that we are given a set of positive, ambiguous, and negative samples $\{(x_i,y_i)\}_{i=1}^{N} $
drawn independently from the probability distribution with density $p_0(x,y)$ defined on $\mathcal{X}\times\mathcal{Y}_0$.
Our goal is still to learn a discriminant function that classifies test samples into either the positive or negative class (not in the ambiguous class).
Our key question in this scenario is if we can utilize the ambiguous training data to improve the classification accuracy of the discriminant function.

In this section, we develop a new method based on a method of \emph{classification with reject option} (CRO)
\citep{cortes2016learning}. For this reason, before deriving the new method, we first review the CRO method.

\subsection{Classification with reject option by SVM (CRO-SVM)}
\cite{cortes2016learning} introduced a rejection function $r:\mathcal{X}\rightarrow\mathbb{R}$ to identify the regions with high risk for misclassification, in addition to discriminating the positive and negative classes. When the rejection function takes a positive value, the corresponding sample is accepted and is classified into the positive or negative class by classifier $h$; otherwise, the sample is rejected and is not classified. When a sample is rejected, the rejection cost $c$ is incurred, which trades off the risk of misclassification and the cost of rejection. To realize this idea, the \emph{0-1-c loss} was introduced:
\begin{equation}
L_\mathrm{01c}(h,r,x,y) = 1_{yh(x)\leq0}1_{r(x)>0}+c1_{r(x)\leq0}. \label{L_0-1-c}
\end{equation}

When $c=0$, all samples are rejected because the loss function does not incur any cost. On the other hand, when $c\geq0.5$, no samples are rejected because the expectation of the 0-1 loss is less than $0.5$; in that case, the 0-1-$c$ loss is reduced to the 0-1 loss. Therefore, effectively, we only consider $c$ such that $0<c<0.5$. The rejection function and the discriminant function are simultaneously learned from training data.

%Corresponding empirical loss is
%\begin{equation}
%R_{0-1-c}(h)=\frac{1}{N}\sum_{i=0}^{N-1}L_{0-1-c}(h,r,x_i,y_i).
%\end{equation}
%の最小化により求めたいが、前節と同様に、$L_{0-1-c}$の上界で凸の代理損失
Similarly to the 0-1 loss, the 0-1-$c$ loss has discrete nature and thus its direct optimization is computationally intractable.
To avoid the discontinuity, the following surrogate loss called the \emph{max-hinge (MH) loss}
was introduced:
\begin{equation}
L_\mathrm{MH}(h,r,x,y) = \max\left(1+\frac{\alpha}{2}\left(r(x)-yh(x)\right),c\left(1-\beta r(x)\right),0\right),
\end{equation}
where $\alpha, \beta>0$ are the hyperparameters to control the shape of the surrogate loss.
% In the same manner, we formulate a empirical risk minimization problem
%\begin{eqnarray}
%(h^*, r^*)=\argmin_{(h,r)}\hat{R}_\mathrm{MH}(h,r)\nonumber \\
%\hat{R}_\mathrm{MH}(h,r) = \frac{1}{N}\sum_{i=1}^{N}L_\mathrm{MH}(h,r,x_i,y_i).
%\end{eqnarray}
%We also parameterize the rejection function using a positive-definite kernel like Eq. (\ref{kernel}),
%\begin{equation}\label{kernel_r}
%r(x;u)=\sum_{i=1}^{N}u_iK(x,x_i).
%\end{equation}
%We introduce an L2 regularization term and reduce it to a quadratic programming problem

In the same manner as the original SVM, introducing the L2 regularization, basis functions, and slack variables yields the following quadratic program:
\begin{eqnarray}
(\hat{w}, \hat{u}, \hat{\xi}) &=& \argmin_{(w, u, \xi)}\left[\frac{\lambda}{2} \| w \|^2 + \frac{\lambda'}{2} \| u\|^2 + \frac{1}{N} \sum_{i=1}^{N}\xi_i \right] \nonumber \\
&&\mathrm{s.t.} \left(
\begin{array}{c}
\xi_i \geq 1 + \frac{\alpha}{2}\left(r_i - y_i h_i \right) \\
\xi_i \geq c(1-\beta r_i) \\
\xi_i \geq 0
\end{array}
\right) \mathrm{for}~i=1,\ldots,N,
\label{h_i}
\end{eqnarray}
where $w=(w_1,\ldots,w_N)^\top$ are the coefficients of the discriminant function,
$u=(u_1,\ldots,u_N)^\top$ are the coefficients of the rejection function,
$\lambda,\lambda'>0$ are the L2 regularization parameters, and
 $h_i$ and $r_i$ denote the values of the discriminant function and rejection function at sample point $x_i$ given by $h_i = \sum_{j=1}^{N}w_j \phi_j(x_i)$ and $r_i=\sum_{j=1}^{N}u_j \phi_j(x_i)$, respectively.
The resulting discriminant function and rejection function are given by
$h(x; \hat{w}) = \sum_{j=1}^N \hat{w}_j \phi_j(x)$
and $r(x; \hat{u}) = \sum_{j=1}^N \hat{u}_j \phi_j(x)$.

We refer to this method as CRO-SVM.

%Originally, this method was proposed to classify the test data to positive, negative and rejected by the criteria below:
%\begin{equation} \label{infer1}
%\left\{
%\begin{array}{ll}
%\mathrm{Positive} &\Leftrightarrow h^*>0, r^*>0 \\
%\mathrm{Negative} &\Leftrightarrow h^*<0, r^*>0 \\
%\mathrm{Rejected} &\Leftrightarrow r^*\leq0
%\end{array}
%\right. 
%\end{equation}
%However, it is difficult to evaluate the model performance since the dataset has only positive and negative labels. Therefore, in this study, we only apply the discriminant function to the test data and classify it to positive and negative like
%\begin{equation} \label{infer2}
%\left\{
%\begin{array}{ll}
%\mathrm{Positive} &\Leftrightarrow h^*>0\\
%\mathrm{Negative} &\Leftrightarrow h^*<0\\
%\end{array}
%\right. 
%\end{equation}
%Note that the binary classification result is different from that of SVM, because training data hard to discriminate are not affect the discriminant function. Binary classification using this method is sometimes superior to SVM, since it focuses on the regions that are relatively easy to discriminate.

\subsection{Proposed method: classification with ambiguous data by SVM (CAD-SVM)}
To handle ambiguous training data in the SVM formulation,
we extend the 0-1-$c$ loss to the \emph{0-1-c-d loss} defined as
\begin{equation}
L_\mathrm{01cd}(h,r,x,y) = 1_{y^2=1}\left(1_{yh(x)\leq0}1_{r(x)>0}+c1_{r(x)\leq0}\right) + d 1_{y=0} 1_{r(x)>0}. \label{L_0-1-c-d}
\end{equation}
Table~\ref{fig_0-1-c_loss} and Table~\ref{fig_0-1-c-d_loss} compare the behavior of the 0-1-$c$ loss and the 0-1-$c$-$d$ loss.
For positive and negative samples, the 0-1-$c$-$d$ loss behaves the same as the 0-1-$c$ loss.
On the other hand, for ambiguous samples, the 0-1-$c$-$d$ loss incurs penalty $d$ when
they are classified into the positive or negative class.
Therefore, ambiguous samples tend to be classified into the ambiguous class
if we employ the 0-1-$c$-$d$ loss.
Compared to the CRO formulation, where a rejector cannot be learned explicitly from positive and negative samples,
the CAD utilizes ambiguous samples to learn a rejector explicitly.

The above discussion may mislead us as if
we are just solving a 3-class problem with the positive, ambiguous, and negative classes.
However, we do not classify test samples into the ambiguous class, but only into the positive and negative classes.
To solve the CAD problem, we utilize a binary discriminant function $h$ and a rejection function $r$,
as in the CRO formulation reviewed above.
More specifically, we train $h$ and $r$ under the 0-1-$c$-$d$ loss,
and we only use $h$ in the test phase to classify test samples into the positive and negative classes.
Thanks to the interplay between $h$ and $r$ in the 0-1-$c$-$d$ loss,
we can utilize ambiguous data to train $h$ through $r$.

\begin{table}
\centering
\caption{The 0-1-$c$ loss function.} 
\scriptsize
\begin{tabular}{|l|c|c|c|}
\hline
\diagbox{Label $y$}{Judgement $(h,r)$} & \begin{tabular}{c}Positive\\$h>0$\\$r>0$\end{tabular} & \begin{tabular}{c}Rejected \\$ r\leq 0 $\end{tabular}& \begin{tabular}{c}Negative \\$ h \leq 0$ \\ $ r > 0 $\end{tabular} \\
\hline
Positive $\quad y=1$ & 0 & $c$ & 1 \\
\hline
Negative $\quad y=-1$ & 1 & $c$ & 0 \\
\hline
\end{tabular}\label{fig_0-1-c_loss}
% \end{table}
% \begin{table}
% \centering
\vspace*{5mm}
\caption{The 0-1-$c$-$d$ loss function.} 
\scriptsize
\begin{tabular}{|l|c|c|c|}
\hline
\diagbox{Label $y$}{Judgement $(h,r)$} & \begin{tabular}{c}Positive\\$h>0$\\$r>0$\end{tabular} & \begin{tabular}{c}Ambiguous \\$ r\leq 0 $\end{tabular}& \begin{tabular}{c}Negative \\$ h \leq 0$ \\ $ r > 0 $\end{tabular} \\
\hline
Positive $\quad y=1$ & 0 & $c$ & 1 \\
\hline
Ambiguous $\quad y=0$ & $d$ & 0 & $d$ \\
\hline
Negative $\quad y=-1$ & 1 & $c$ & 0 \\
\hline
\end{tabular}\label{fig_0-1-c-d_loss}
\end{table}
%\begin{table}
%\centering
%\caption{The schematic image of the 0-1-c-d loss.} 
%\begin{tabular}{|c|c|c|c|}
%\hline
%\diagbox{Label}{Judgement} & \begin{tabular}{c}Positive\\$h>0$\\$r>0$\end{tabular} & \begin{tabular}{c}Ambiguous \\$ r\leq 0 $\end{tabular}& \begin{tabular}{c}Negative \\$ h < 0$ \\ $ r > 0 $\end{tabular} \\
%\hline
%Positive & 0 & $c$ & 1 \\
%\hline
%Ambiguous & $d$ & 0 & $d$ \\
%\hline
%Negative & 1 & $c$ & 0 \\
%\hline
%\end{tabular}\label{fig_0-1-c-d_loss}
%\end{table}

%これの経験損失
%\begin{equation}
%R_{0-1-c-d}(h,r) = \frac{1}{N} \sum_{i=0}^{N-1} L_{0-1-c-d}(h,r,x_i,y_i)
%\end{equation}
%を最小化することを考える。

\begin{figure}
\centering
\includegraphics[width=10cm,bb=0 0 800 550]{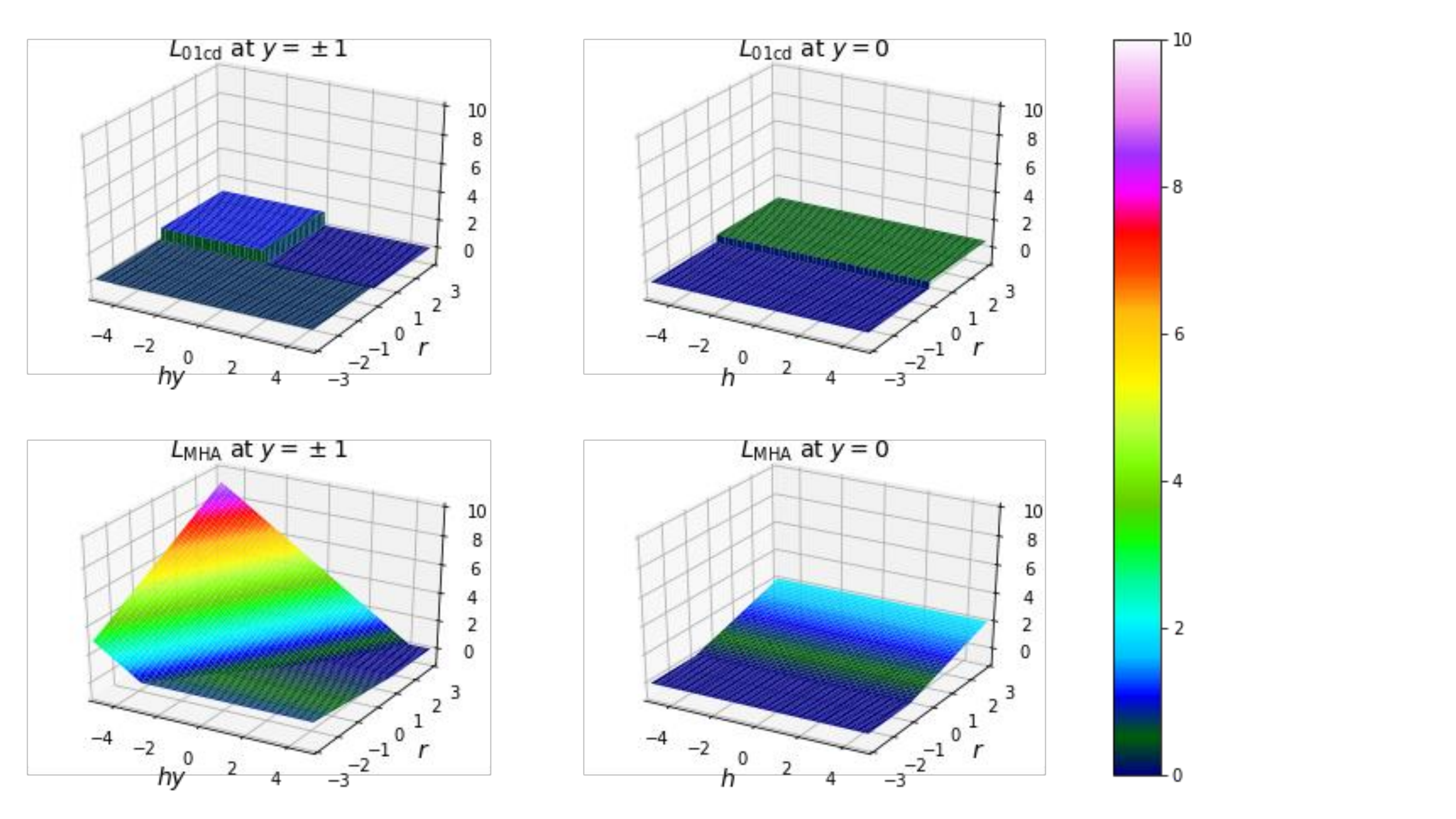}
\caption{The 0-1-$c$-$d$ loss $L_\mathrm{01cd}$ and its surrogate loss $L_\mathrm{MHA}$ for the penalty values $(c, d) = (0.2, 0.5)$.}
\label{fig7}
\end{figure}

Similarly to the 0-1-$c$ loss, we consider the following convex upper bound
of the 0-1-$c$-$d$ loss
called the \emph{max-hinge-ambiguous (MHA) loss}
as a surrogate to avoid its discrete nature:
\begin{eqnarray}
%L_{0-1-c-d}(h,r,x,y) \leq 1_{y^2=1} L_{MH}(h,r,x,y) + d 1_{y=0} \max\left(1+\gamma r(x), 0\right) \nonumber \\
%=y^2 L_{MH}(h,r,x,y) + (1-y^2) \max \left(d\left(1+\gamma r(x)\right), 0\right) \nonumber \\
%\leq y^2 L_{MH}(h,r,x,y) + (1-y^2) \max \left(\eta d\left(1+\gamma r(x)\right), 0\right) \nonumber \\
%= y^2 \max \left( 1 + \frac{\alpha}{2}\left(r(x) - yh(x)\right), c\left(1 - \beta r(x)\right), 0 \right) \nonumber \\
% + (1-y^2) \max \left(\eta d\left(1+\gamma r(x)\right), 0\right)
%\equiv L_{MHA} (h,r,x,y)
L_\mathrm{01cd}(h,r,x,y) &\leq& 1_{y^2=1} L_\mathrm{MH}(h,r,x,y) + d 1_{y=0} \max\left(1+\beta r(x), 0\right) \nonumber \\
&=& y^2 \max \left( 1 + \frac{\alpha}{2}\left(r(x) - yh(x)\right), c\left(1 - \beta r(x)\right), 0 \right) \nonumber \\
&& + (1-y^2) \max \left(d\left(1+\beta r(x)\right), 0\right) \nonumber \\
&\leq& y^2 \max \left( 1 + \frac{\alpha}{2}\left(r(x) - yh(x)\right), \eta c\left(1 - \beta r(x)\right), 0 \right) \nonumber \\
&& + (1-y^2) \max \left(\eta d\left(1+\beta r(x)\right), 0\right) \nonumber \\
&\equiv& L_\mathrm{MHA} (h,r,x,y),
\end{eqnarray}
where $\eta\geq 1$ is the hyperparameter
to control the shape of the surrogate loss.
See Figure~\ref{fig7} for its visualization.

Then, in the same way as the CRO-SVM, we have the following quadratic program:
%\begin{equation}
%\hat{R}_\mathrm{MHA}(h,r) = \frac{1}{N} \sum_{i=1}^{N} L_\mathrm{MHA}(h,r,x_i,y_i).
%\end{equation}
%
%We express $(h, r)$ using kernels like Eqs. (\ref{kernel},\ref{kernel_r}) and add L2 regularization terms. Then, 
\begin{eqnarray}
(\hat{w}, \hat{u}, \hat{\xi})&=& \argmin_{(w,u,\xi)}\left[ \frac{\lambda}{2} \|w \|^2 + \frac{\lambda'}{2} \| u \|^2 + \frac{1}{N}\sum_{i=1}^{N}\xi_i \right] \nonumber \\
&&\mathrm{s.t.} \left(
\begin{array}{c}
\xi_i \geq y_i^2\left(1+\frac{\alpha}{2}(r_i - y_i h_i)\right) \\
\xi_i \geq y_i^2 \eta c(1-\beta r_i) \\
\xi_i \geq (1-y_i^2) \eta d(1+\beta r_i)
\end{array}
\right) \mathrm{for}~i=1,\ldots,N.
\end{eqnarray}
This is our proposed method called CAD-SVM.
% 20200726 added
The computational complexity of the CAD-SVM depends on implementation of the quadratic program. It naively costs $O(N^3)$, but if we use a fixed number of basis functions, the complexity reduces to $O(N)$.
% end 20200726

% \subsection{Choice of hyperparameter $\eta$}

%  Theorem 1 suggests the hyperparameters that we can obtain hyperparameters that give a good approximation of the 0-1-c-d loss.

% In this study, we define the approximation based on the Bayesian classifiers. Generally, the Bayesian classifiers are defined to be

% For each $x\in\mathcal{X}$ , let
% \begin{equation}
% \left(h^*(x), r^*(x)\right) = \argmin_{(h,r)} \mathbb{E}_{y\sim p_0(y|x)} \left[ l(h,r,x,y) \right],
% \end{equation}
% where $l(h,r,x,y)$ is an arbitrary loss function

% . Since the predicted label of the sample is determined by the signs of $h$ and $r$, if the signs of the Bayesian classifiers of the two loss functions match for all conditional probabilities $p_0(y|x)$, one can be considered as a good approximation of another loss function.
% % If the probabilistic distribution $p_0(x,y)$ is given, the optimal values of $h(x)$ and $r(x)$ are determined by the value of the loss functions and the probability of the label $p_0(y|x)$. The optimal values of $(h, r)$ subject to the conditional probability $p_0(y|x)$ is called Bayesian classifiers. Since the classification is determined by the signs of $h(x)$ and $r(x)$, if the signs of the Bayesian classifiers of the two loss functions match for all conditional probabilities, one can be considered as a good approximation of another loss function.

The MHA loss depends on the choice of hyperparameters $(\alpha, \beta, \eta)$.
To find good hyperparameter values, let us analyze the 0-1-$c$-$d$ loss first.

\begin{figure}
\centering
\includegraphics[width=5cm,bb=0 200 400 550]{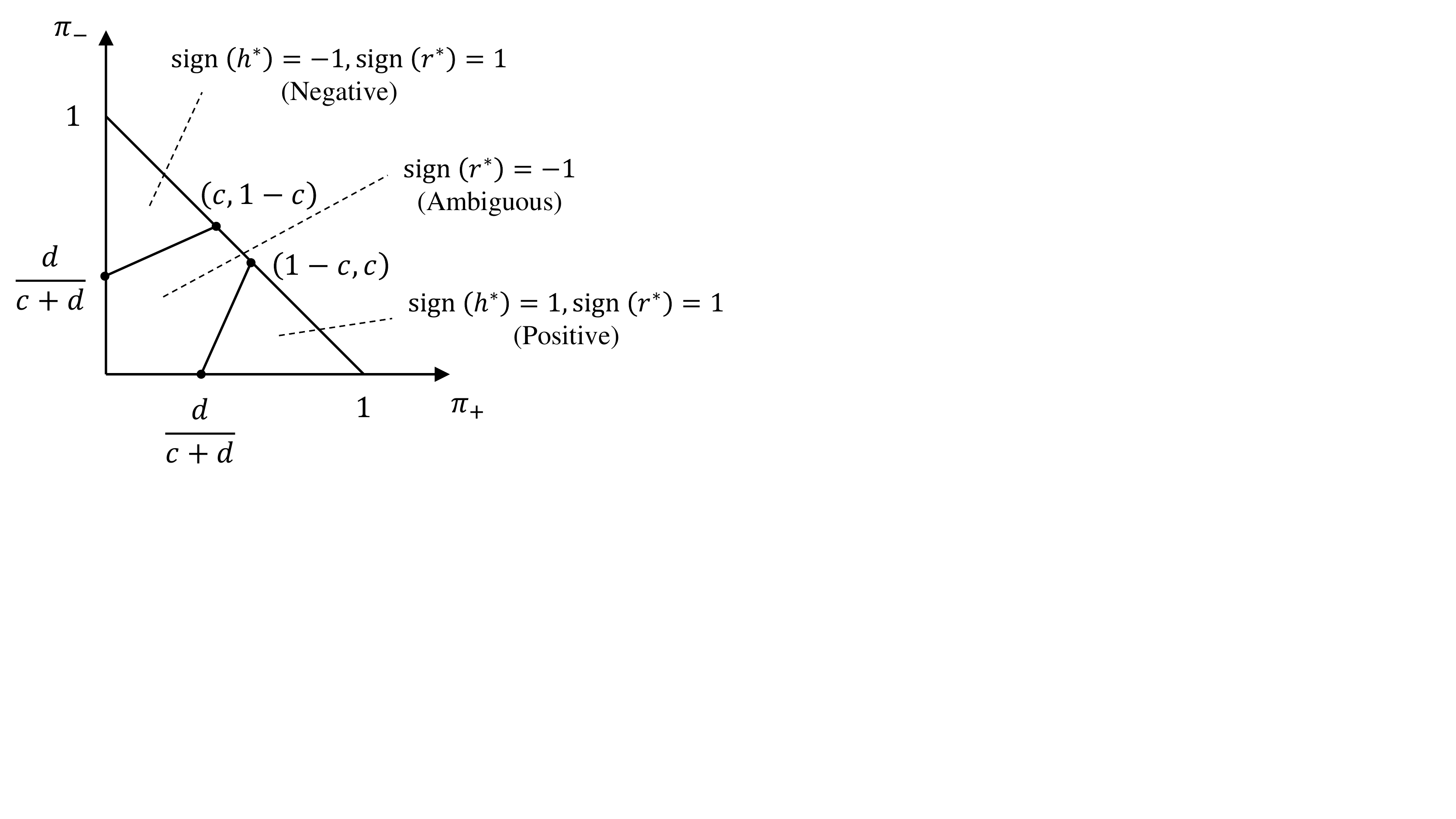}
\caption{Optimal solutions for the 0-1-$c$-$d$ loss.}
\label{fig1}
\end{figure}

For each $x\in\mathcal{X}$, let $\pi_+(x) = p_0(y=1|x)$, $\pi_0(x) = p_0(y=0|x)$, and $\pi_-(x) = p_0(y=-1|x)$, where $\pi_+(x) +\pi_0(x) + \pi_-(x) = 1$.
Then the following lemma shows how $c$ and $d$ are related to $\pi_+(x)$ and $\pi_-(x)$ for the optimal classifier and rejector (its proof is available in Appendix~\ref{sec:proof-lemma}):
\begin{lemma} \label{lem1}
For each $x\in\mathcal{X}$, let
\begin{equation}
\left(h_\mathrm{01cd}^*, r_\mathrm{01cd}^*\right)
= \argmin_{(h,r)} \mathbb{E}_{p_0(y|x)} \left[ L_\mathrm{01cd}(h,r,x,y) \right].
\end{equation}
Then
\begin{equation}
\left\{
\begin{array}{ll}
\sign(h^*_\mathrm{01cd})=1,~\sign(r^*_\mathrm{01cd})=1 & \mathrm{if}~\pi_+\geq\frac{\disp d+(1-c-d)\pi_-}{\disp c+d}, \\
\sign(h^*_\mathrm{01cd})=-1,~\sign(r^*_\mathrm{01cd})=1 & \mathrm{if}~\pi_-\geq\frac{\disp d+(1-c-d)\pi_+}{\disp c+d}, \\
\sign(r^*_\mathrm{01cd})=-1& \mathrm{otherwise}.
\end{array}
\right.
\end{equation}
%and the expectation value of the 0-1-c-d loss is
%\begin{equation}
%\mathbb{E}_{y\sim p_0(y|x)}[L_\mathrm{01cd}(h,r,x,y)]=
%\left\{
%\begin{array}{ll}
%d\pi_0+\pi_- & \left(\pi_+\geq\frac{\disp d+(1-c-d)\pi_-}{\disp c+d}\right) \\
%d\pi_0+\pi_+ & \left(\pi_-\geq\frac{\disp d+(1-c-d)\pi_+}{\disp c+d}\right) \\
%c(\pi_+ + \pi_-)& (\mathrm{otherwise})
%\end{array}
%\right.
%\end{equation}
\end{lemma}
Figure~\ref{fig1} illustrates the above results.
This shows that when $\pi_+$ (or $\pi_-$) is large (i.e., imbalanced classification),
the rejector accepts the sample and
the classifier classifies that sample into the positive (or negative) class.
On the other hand, when both $\pi_+$ and $\pi_-$ are not so large,
the rejector rejects the sample.

Next, based on the above lemma, we have the following theorem for the MHA loss
(its proof is given in Appendix~\ref{sec:proof-theorem}):
\begin{theorem} \label{theo1}
For each $x\in\mathcal{X}$, let
\begin{equation}
\left(h_\mathrm{MHA}^*, r_\mathrm{MHA}^*\right)
= \argmin_{(h,r)} \mathbb{E}_{p_0(y|x)} \left[ L_\mathrm{MHA}(h,r,x,y) \right].
\end{equation}
Then, for 
\begin{equation}
% \alpha = 1, \quad\beta = \frac{1}{1-2c}, \quad\gamma = \frac{1}{2(1-c)(1-2c)}, \quad\eta=2(1-c), \label{theo1eq1}
\alpha^* = 2(1-2c), \quad\beta^* = 1+2c, \quad\eta^*=\frac{2}{1+2c}, \label{theo1eq1}
\end{equation}
the signs of $(h^*_\mathrm{MHA}, r^*_\mathrm{MHA})$ match those of $(h^*_\mathrm{01cd},r^*_\mathrm{01cd})$.
\end{theorem}

Based on the above theorem, 
we use Eq.~(\ref{theo1eq1}) as hyperparameter values in our experiments in the next section
and demonstrate that they work well in practice.
Nevertheless, given that the above theorem is valid only for the optimal solutions,
we may cross-validate better hyperparameter values around Eq.~(\ref{theo1eq1})
to further improve the classification performance.
Note that Eq.~(\ref{theo1eq1}) does not include $d$.

\section{Numerical experiments} \label{Num}
In this section, we report experimental results.

\subsection{Datasets}
For experiments, we use a toy dataset, a public dataset, and an in-house dataset.

\subsubsection{Toy dataset}
\begin{figure}[t]
\centering
\includegraphics[width=5cm,bb=0 150 400 550]{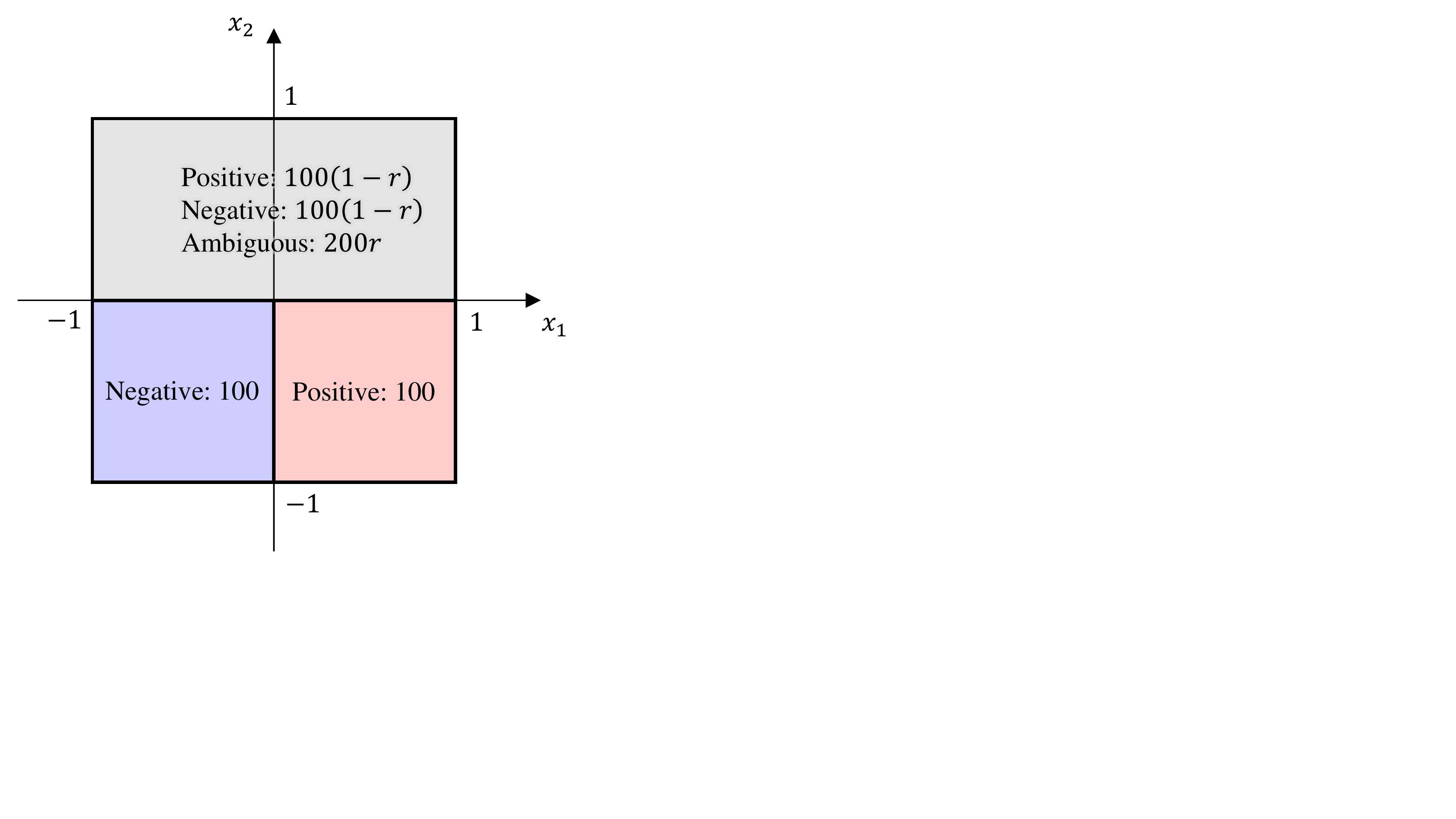}
\caption{Toy dataset. The lower-left and lower-right regions are negative and positive regions, while the upper region is the mixed region. 
The numbers indicate the numbers of samples in each region. For each region, samples are distributed uniformly.}
\label{fig_MM}
\end{figure}

To understand the behavior of our method and related methods, we created a toy classification problem and applied the methods to it. The problem contains three regions: the positive, negative, and mixed regions (see Figure~\ref{fig_MM}). The positive and negative regions are clearly separable, whereas the mixed region has no good discriminant function. For this problem, we want to clearly discriminate the positive and negative regions, with the influence of the mixed region avoided.
%We studied how the proportion of ambiguous samples influence the results by changing the parameters $r_1, r_2$, where $r_1$ corresponds to the ratio of the mixed region and $r_2$ corresponds to the ratio of ambiguous samples in the mixed region. The number of all samples are 400, the total number of positive and negative samples in the separable regions are $400 (1-r_1)$, and the total number of positive and negative samples in the mixed region is $400 r_1 (1-r_2)$. Therefore, the expected maximum accuracy is $\{(1-r_1) + r_1 (1-r_2) \times0.5\}/\{(1-r_1) + r_1 (1-r_2)\}$.
We also studied how the proportion of ambiguous samples influences the results by changing the proportion of ambiguous samples in the mixed region $r$. The number of all samples is 400, the total number of positive and negative samples in the separable regions is $200$, and the total number of positive and negative samples in the mixed region is $200 (1-r)$. Therefore, the expected maximum accuracy is $\{1 + (1-r) \times0.5\}/\{1 + (1-r)\}$.

\subsubsection{Public datasets  (PD1, PD2, PD3)}
% As far as we know, there are no public dataset that contains ambiguous labels. It is not appropriate that we manually select ambiguous samples from a classification dataset or that we select ambiguous samples by the classification with reject option. Therefore, 
We processed a regression dataset, the Boston Housing Dataset \citep{harrison1978hedonic}, to convert it to a classification dataset with ambiguous data.
The original dataset consists of 13 features $\xi_i\in\mathbb{R}^{13}$ associated with the average house prices $\zeta_i\in\mathbb{R}$ for 506 districts, where $1\leq i \leq 506$ denotes the sample number. We annotated all the samples to be positive, negative, and ambiguous according to the following procedure:

\begin{description}
\item[PD1 (P/N/A separable):] Simply, the samples with $\zeta_i > 23$ were labeled as positive, the samples with $\zeta_i < 19$ were labeled as negative, and the other samples were labeled as ambiguous. The numbers of samples were 190, 173, and 143 for the positive, negative, and ambiguous classes, respectively.
\item[PD2 (P/N/A mixed):] The samples with $\zeta_i > 23$ and the samples with $\zeta_i < 19$ were still labeled as positive and negative, respectively, but the remaining 143 samples were randomly labeled as positive, negative, or ambiguous.
%\item[PD1:] Simply, the districts whose average house prices are over \$23,000 were labeled as positive, under \$19,000 were labeled as negative, and between \$19,000 and \$23,000 were labeled as ambiguous. The numbers of samples are 190, 173, and 143 for the positive, negative, and ambiguous classes, respectively.
%\item[PD2:] More realistically, the districts with an average house price over \$23,000 and under \$19,000 were still labeled as positive and negative, respectively, but the remaining 143 samples were randomly labeled as positive, negative, or ambiguous.
\item[PD3 (separable and mixed):] We considered a hyperplane $v\cdot\xi=0$ in the feature space and divided all the samples into two parts, the mixed part $\{i|v\cdot\xi_i\geq0\}$ and the separable part $\{i|v\cdot\xi_i<0\}$. The coefficients of the hyperplane $v$ are selected so that the averages of $\zeta_i$ over the samples in both parts were approximately matched. For the mixed part, each sample is randomly labeled as positive, negative or ambiguous, whereas for the separable part, the samples with $\zeta_i > 21$ were labeled as positive and the others were labeled as negative. The numbers of samples in the separable region were 83, 0, and 87, and those in the mixed region were 114, 107, and 115 for the positive, negative, and ambiguous classes, respectively.
\end{description}

Figure~\ref{figs_PCA} (a)--(c) and (e)--(g) show the 2-dimensional plots of the above datasets
visualized by the principle component analysis (PCA) and the locality preserving projection (LPP) \citep{NIPS16:He+Niyogi:2004}. On the whole, ambiguous points are located between positive and negative points.

\subsubsection{In-house dataset from a cell culture process (ID)}
As a real-world application, we prepared an in-house cell culture dataset. This dataset contained 124 fields of view (FOV). For each FOV, 2 images were acquired: one was in the middle, and the other was at the end of the culturing process. Each middle image was analyzed by the image processing software, CL-Quant \citep{alworth2010teachable}, and converted to 8 morphological features such as the average brightness and average area of cells.

Each final image was annotated by experts. If the cells in the image overall looked healthy/damaged, the image was labeled as positive/negative. However, some images contained both healthy and damaged cells, and they were labeled as ambiguous. The numbers of samples were 41, 59, and 24 for positive, negative, and ambiguous, respectively.

% added revised version 2020/06/29
Our motivation was to predict the final state of each FOV annotated by the experts, using morphological features obtained in the middle of the culturing process.
If we could predict it accurately, the culturing cost would be saved by aborting the culturing process where the cells would be damaged while keeping the healthy cells cultured.
Therefore, we should focus on reducing misclassification between the positive and negative classes and we hope that information from ambiguous samples could be useful to improve the prediction accuracy.
That was why test samples did not have the ambiguous label and were not classified into the ambiguous class in our scenario.
Though ambiguous samples may occur in the test phase in the actual curturing process, we did not care which class they were classified into.
% end revision

\begin{figure}
\centering
\includegraphics[width=13cm,bb=0 180 1000 550]{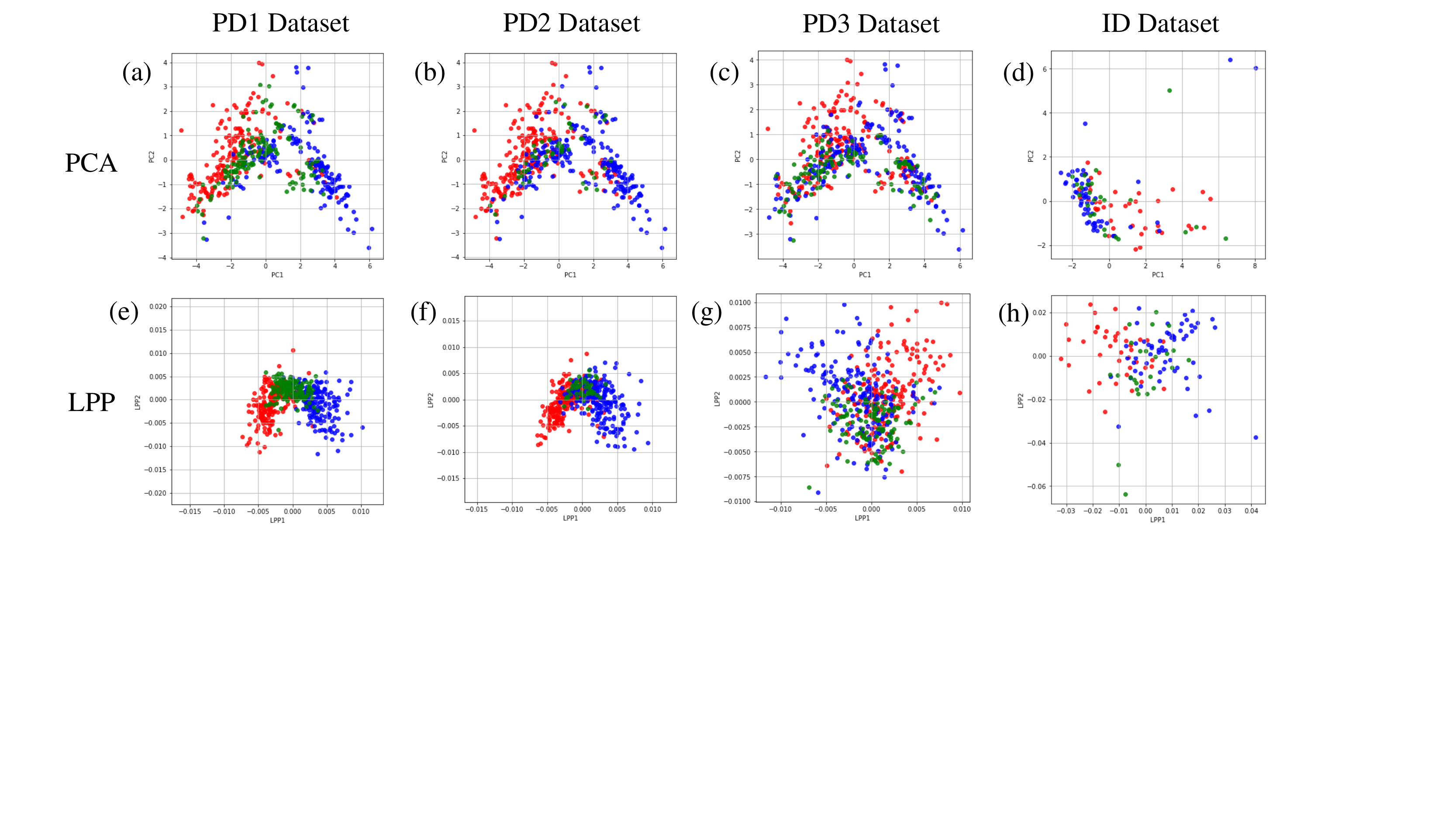}
\caption{Two-dimensional visualization of the PD1, PD2, PD3, and ID datasets by the PCA and the LPP. Red, blue, and green points correspond to positive, negative and ambiguous samples, respectively.}
\label{figs_PCA}
\end{figure}

In the same manner as the PD1, PD2, and PD3 datasets, this dataset was also visualized by the PCA and the LPP in Figure~\ref{figs_PCA} (d) and (h), respectively.
They show that ambiguous points are roughly located between positive and negative points.

\subsection{Experimental settings}
Using the above datasets, we compared the classification performance of the SVM, the SVM-RL (random label), the LapSVM \citep{belkin2006manifold}, the two-step SVM, the CRO-SVM, the CRO-SVM-RL, and the CAD-SVM. The SVM-RL employs the SVM algorithm, but ambiguous samples are randomly relabeled as positive or negative, which effectively utilizes information of the ambiguous samples. The LapSVM is a semi-supervised learning method based on the SVM, which employs a regularization term defined by the graph Laplacian. Ambiguous samples are treated as unlabeled samples in the LapSVM. The two-step SVM is the method which learns the rejection function and the discriminant function sequentially---first the rejection function is learned to judges whether the sample is ambiguous or not, and then the discriminant function is learned only using the samples which are not rejected by the rejection function. When the rejection function is learned, class weights $c$ and $d$ are applied to non-ambiguous (i.e., positive and negative) and ambiguous samples, respectively. The CRO-SVM-RL is the CRO-SVM with random labels for ambiguous samples in the same manner as SVM-RL.
%Table~\ref{tab_method_comparison} shows the qualitative comparison of those methods.

%\begin{table}[] \centering
%\caption{The qualitative differences among the baseline methods in the numerical experiments. The acronyms P, N, A and U denote Positive, Negative, Ambiguous and Unlabeled, respectively.}
%\scriptsize
%\begin{tabular}{lll} \hline
%% & P/N samples & A samples \\ \hline
%%SVM & Treated as P/N & Ignored \\ \hline
%%SVM-RL & Treated as P/N & Randomly labeled to P or N \\ \hline
%%LapSVM & Treated as P/N & Treated as U \\ \hline
%%Two-step SVM & Treated as P/N & Treated as A \\ \hline
%%CRO-SVM & Treated as P/N or A & Ignored \\ \hline
%%CRO-SVM-RL & Treated as P/N or A & Randomly labeled to P or N \\ \hline
%%CAD-SVM & Treated as P/N or A & Treated as A \\ \hline
%&Leanrt models &A samples \\ \hline \hline
%SVM & Classifier only &  Ignored \\ \hline
%SVM-RL & Classifier only & Randomly labeled to P or N \\ \hline
%LapSVM & Classifier only & Treated as U \\ \hline
%Two-step SVM & Classifier and rejector & Treated as A \\ \hline
%CRO-SVM & Classifier and rejector & Ignored \\ \hline
%CRO-SVM-RL & Classifier and rejector & Randomly labeled to P or N \\ \hline
%CAD-SVM & Classifier and rejector &Treated as A \\ \hline
%
%\end{tabular}
%\label{tab_method_comparison}
%\end{table}

For each method, 500 test runs were performed by changing the training and test datasets which were randomly divided from the original dataset. The dividing ratio of the training and test dataset was 4:1 for the ID dataset and 1:2 for the other datasets. For each test run, 5-fold cross validation was performed to determine the parameters below. For validation and in the test phase, only positive and negative samples were applied to the discriminant function, thus we were able to evaluate the binary classification accuracy.

% additional information 7/22
Note that our goal was to maximize the binary classification accuracy in the test phase, and that was equivalent to minimizing the expected 0-1 loss.
Though we trained the models using various loss functions such as the hinge loss, the MH loss, and the MHA loss, the 0-1 loss was minimized by cross validation.
%The two-step SVM, the CRO-SVM, the CRO-SVM-RL, and CAD-SVM contain $c$ and $d$, and we tested multiple patterns for $c$ and $d$ in the experiments.

In total, we had 10 hyperparameters $(\lambda, \lambda', \sigma, \sigma', \tau, c, d, \alpha, \beta, \eta)$, where $(\lambda, \lambda')$ were the L2 regularization parameters, $\sigma$ was the width of the Gaussian radial basis function in the basis functions $\phi_i(x) = \exp\left( -\frac{\| x-x_i\|^2}{2\sigma^2}\right)$, $\sigma'$ was the hyperparameter of the weight matrix $W$ of the graph Laplacian as $W_{ij} = \exp\left( - \frac{\|x_i-x_j\|^2}{2\sigma'^2}\right)$ (only in the LapSVM), $\tau$ was the coefficient of the graph Laplacian regularization (only in the LapSVM), $(c, d)$ were the hyperparameters of the 0-1-$c$ loss (in the CRO-SVM and CRO-SVM-RL), the 0-1-$c$-$d$ loss (in the CAD-SVM) or the class weights (in the two-step SVM), and $(\alpha, \beta, \eta)$ were the hyperparameters of the MH and MHA loss functions (in the CRO-SVM, CRO-SVM-RL and CAD-SVM). We used 5-fold cross validation in terms of the classification accuracy to choose the hyperparameters from $(\lambda, \lambda')\in\{10^{-3}, 10^{-5}, 10^{-7}\}$, $(\sigma, \sigma') \in \{10^{0.5}, 10^{0.75}, 10^1\}$, $\tau\in\{10^{-1}, 10^{-2}, 10^{-3}\}$, $c \in \{0.03, 0.06, 0.20, 0.45\}$, and $d\in\{0.03, 0.06, 0.20, 0.50\}$. The other hyperparameters $(\alpha, \beta, \eta)$ were determined by Eq.~(\ref{theo1eq1}). Quadratic programming problems were solved using cvxopt \citep{cvxopt}.

\subsection{Results}

\begin{figure}
\centering
\includegraphics[width=11cm,bb=0 20 1000 570]{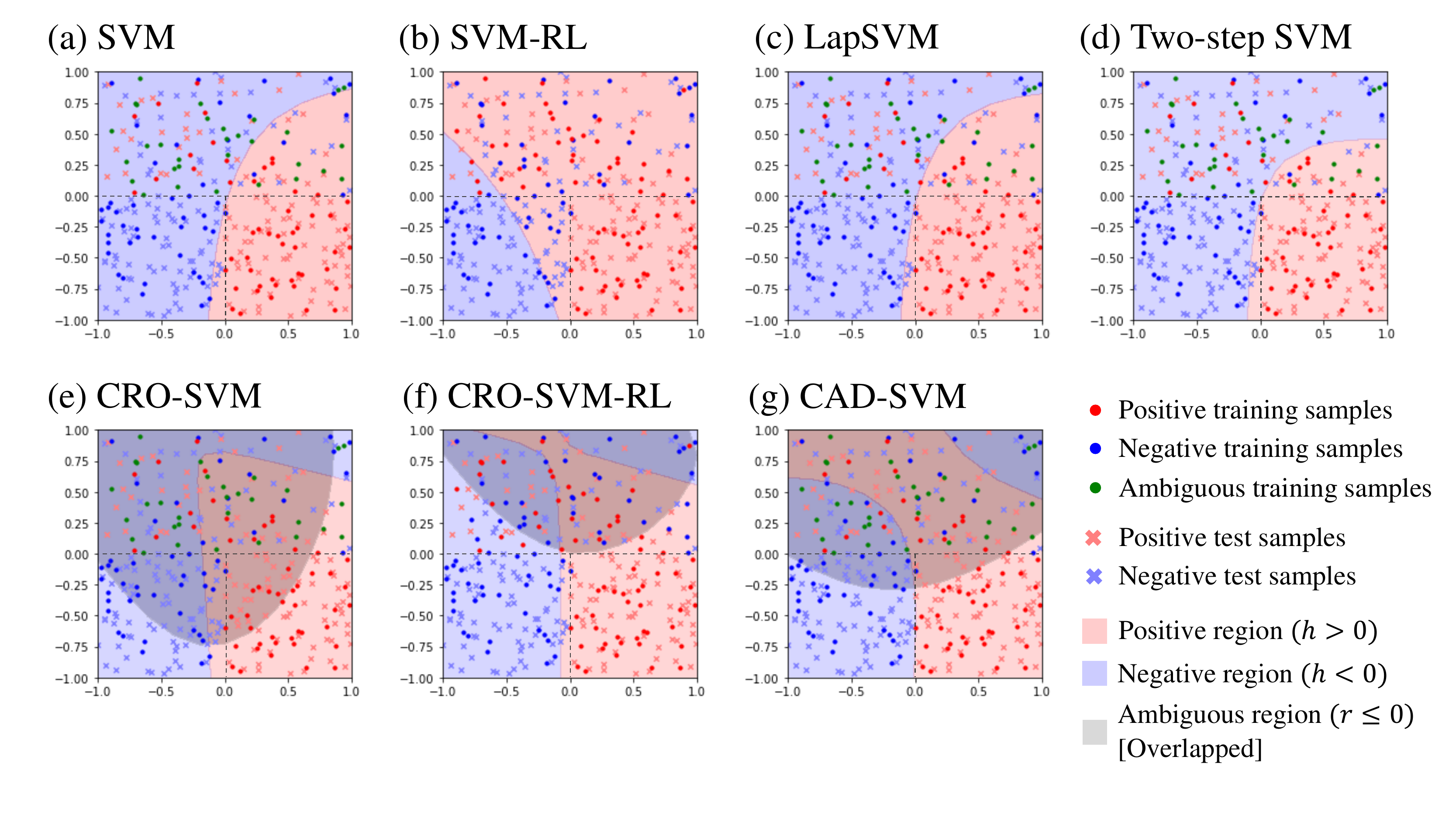}
\caption{An example of discriminant results for the toy dataset.}
\label{figs_MM}
\end{figure}

%Figure~\ref{figs_MM} shows an example of the discriminant results for the toy dataset with $r_1=r_2=0.5$.
Figure~\ref{figs_MM} shows an example of the discriminant results for the toy dataset with $r=0.5$.
The upper half of the domain was the mixed region of positive and negative samples and samples in that region were expected to be classified into the ambiguous class.
On the other hand, the lower half of the domain was perfectly separable into the positive and negative regions and samples in those regions were expected to be discriminated accurately (see Figure~\ref{fig_MM}).
The SVM made some misclassifications in the lower half of the domain, which were caused by the mixed region.
The SVM-RL and LapSVM could not reduce the number of misclassifications, though they utilized information of ambiguous samples.
This suggests that ambiguous samples should not be simply relabeled to positive or negative samples or should not be treated as unlabeled samples.
The two-step SVM could not also reduce the number of misclassifications and it could not learn the ambiguous region.
Thus, it would be disadvantageous to learn the ambiguous region by combining positive and negative classes.
The CRO-SVM and CRO-SVM-RL successfully learned the ambiguous region but that did not lead to learning more accurate discriminant functions.
The CAD-SVM also successfully learned the ambiguous region and then it was able to learn a more accurate discriminant function.
%The CRO-SVM, which identifies the ambiguous region only using positive and negative data, was able to reduce the number of misclassifications, but it was not still the best.
%The two-step SVM and the CAD-SVM successfully learned an accurate discriminant function, meaning that utilizing the information of ambiguous samples can improve classification accuracy.
%But, the two-step SVM could not perfectly identify the mixed region.
Overall, the CAD-SVM was shown to be the most appropriate method in this toy experiment.
%That learned the ambiguous region in the upper half and the middle of the lower half of the domain. The upper ambiguous region would be helpful to learn an accurate discriminant function, while lower ambiguous regions would be harmful. As a result, the number of misclassifications is not still the best, though it is better than the SVM. The CAD-SVM with $c=0.45$ and $d=0.5$ successfully learned the ambiguous region to be only the upper half of the domain, which led to the least misclassifications in the lower half of the domain.
%Therefore, it was needed to reject the upper region not to affect the discriminant function.
%Considering that the SVM is regarded as a special case of the proposed method $(c=0.5, d=0)$ and the CRO is also a special case $(d=0)$, decreasing $c$ and increasing $d$ makes the the rejected region larger. At the condition that $c=0.2$, the CRO learns the rejected region around the upper region and the center of the lower region. The upper rejected region is helpful to learn the more accurate discriminant function, but lower rejected region is harmful. Then, the number of the misclassifications is decreased compared by the SVM, but it is not still the best. On the other hand, the proposed method $(c=0.45, d=0.5)$ learns the rejection region only the upper region, that lead to the least misclassifications in the lower region.

%\hl{Tables}~\ref{tab_toy1} and \ref{tab_toy2} show the test accuracy of toy dataset by changing the parameters $r_1$ and $r_2$ which are related to the propotion of the ambiguous samples.
%As shown in Table~\ref{tab_toy1}, when the ambiguous samples were districuted too widely or too narrowly, the CAD-SVM was not superior to the other methods.
%Table~\ref{tab_toy2} shows when the numbers of the ambiguous samples are small, it also did not work effectively.
%However, when the ambiguous samples distirbuted with the reasonable range and population, the CAD-SVM achieved better score than the other methods, except for the CRO-SVM-RL.
Tables~\ref{tab_toy2} shows the test accuracy of the toy dataset by changing the parameters $r$ which are the propotion of ambiguous samples in the mixed region.
When the proportion of the ambiguous samples was small, the proposed method did not work effectively since the number of ambiguous samples was too small.
On the other hand, when the proportion of the ambiguous samples was large, it also did not show better performance. This was because, under this condition, only small numbers of positive and negative samples existed in the mixed region. Therefore, methods which do not utilize
ambiguous samples showed reasonably good enough performance.
However, when the proportion of ambiguous samples was $r=0.5$, the CAD-SVM achieved a better score than the other methods.

%\begin{table}[] \centering
%\caption{Test accuracy for the toy dataset by changing the ratio of the area of the mixed region $r_1$. The ratio of the ambiguous samples in the mixed region $r_2$ is fixed to 0.5. The boldface numbers show the best and the equivalent results with 5\% t-test.}
%\begin{tabular}{lrrrrr} \hline
%$r_1$&0.1&0.3&0.5&0.7&0.9\\ \hline \hline
%SVM&$\bf{0.958}$&$0.884$&$0.816$&$0.706$&$0.555$\\ \hline
%SVM-RL&$\bf{0.958}$&$0.887$&$0.817$&$0.703$&$\bf{0.558}$\\ \hline
%LapSVM&$0.952$&$0.881$&$0.815$&$0.700$&$0.554$\\ \hline
%Two-step SVM&$0.956$&$0.886$&$0.815$&$0.709$&$\bf{0.560}$\\ \hline
%CRO-SVM&$0.957$&$0.891$&$\bf{0.824}$&$0.714$&$0.556$\\ \hline
%CRO-SVM-RL&$\bf{0.959}$&$\bf{0.893}$&$\bf{0.824}$&$\bf{0.719}$&$0.551$\\ \hline
%CAD-SVM&$0.957$&$\bf{0.891}$&$\bf{0.826}$&$\bf{0.716}$&$\bf{0.563}$\\ \hline
%\end{tabular}
%\label{tab_toy1}
%\end{table}

\begin{table}[] \centering
%\caption{Test accuracy for the toy dataset by changing the ratio of the ambiguous samples in the mixed region $r_2$. The ratio of the area of the mixed region $r_1$ is fixed to 0.5. The boldface numbers show the best and the equivalent results with 5\% t-test.}
\caption{Test accuracy for the toy dataset by changing the ratio $r$ of ambiguous samples in the mixed region. The boldface numbers show the best and equivalent results with 5\% t-test.}
\scriptsize
\begin{tabular}{lrrrrr} \hline
$r$&0.1&0.3&0.5&0.7&0.9\\ \hline \hline
SVM&$\bf{0.739}$&$0.767$&$0.814$&$0.853$&$0.931$\\ \hline
SVM-RL&$\bf{0.739}$&$0.766$&$0.812$&$0.850$&$0.926$\\ \hline
LapSVM&$\bf{0.739}$&$0.766$&$0.814$&$0.850$&$0.930$\\ \hline
Two-step SVM&$\bf{0.737}$&$\bf{0.771}$&$0.816$&$0.854$&$\bf{0.935}$\\ \hline
CRO-SVM&$0.736$&$0.767$&$\bf{0.820}$&$0.857$&$0.931$\\ \hline
CRO-SVM-RL&$0.735$&$0.768$&$0.817$&$\bf{0.861}$&$0.926$\\ \hline
CAD-SVM&$0.736$&$0.767$&$\bf{0.822}$&$0.857$&$0.932$\\ \hline

\end{tabular}
\label{tab_toy2}
\end{table}

Table~\ref{tab_res} summarizes the test accuracy of each method for the PD1, PD2, PD3, and ID datasets, respectively.
For the PD1 dataset, the CAD-SVM was not superior to other methods since this condition was similar to the toy dataset with higher $r$.
For the PD2 dataset, though the dataset had a mixed region, the CAD-SVM also did not show good performance.
However, for the PD3 dataset, which had a mixed region and a separable region, the CAD-SVM achieved the best performance among the compared methods.
It is suggested that our proposed method, the CAD-SVM, works effectively when the dataset has both of a mixed region and a separable boundary between the positive and negative classes.
The CAD-SVM utilizes ambiguous samples to learn a rejector which rejects mixed regions, thus it would be able to focus on learning a separable region.
The CRO-SVM and the CRO-SVM-RL also learn a rejector from positive and negative samples, but the CAD-SVM would be superior since the CAD-SVM can explicitly learn a rejector using ambiguous samples.
For the ID dataset, the CAD-SVM performed better than the other method, except for the CRO-SVM-RL.
Overall, ambiguous samples can improve the binary classification accuracy under some conditions, and the CAD-SVM is one of the solutions that can utilize such ambiguous samples.
%  The parameters $(c, d)$ defined in Eqs.~(\ref{L_0-1-c}) and (\ref{L_0-1-c-d}) correspond to the penalties of misclassification, so they should be determined according to how large ambiguous region is allowed. However,
%Since ambiguous samples were distributed in the mixed region in the PD1, PD2, and ID datasets (see Figure~\ref{figs_PCA}), the CAD-SVM, which identifies the ambiguous region using ambiguous data, was able to ignore the influence of the mixed region and was effective to improve the classification accuracy.

\begin{table}[] \centering
\caption{Test accuracy for the PD1, PD2, PD3, and ID datasets, where $\pm$ denotes the standard deviation. The boldface numbers show the best and equivalent results with 5\% t-test.}
\scriptsize
\begin{tabular}{lrrrr} \hline
&PD1&PD2&PD3&ID\\ \hline \hline
SVM&$\bf{0.924\pm0.019}$&$\bf{0.828\pm0.019}$&$0.635\pm0.029$&$0.803\pm0.089$\\ \hline
SVM-RL&$0.918\pm0.021$&$\bf{0.827\pm0.020}$&$0.629\pm0.030$&$0.806\pm0.088$\\ \hline
LapSVM&$0.921\pm0.020$&$0.824\pm0.021$&$0.629\pm0.031$&$0.802\pm0.088$\\ \hline
Two-step SVM&$0.918\pm0.024$&$0.823\pm0.021$&$0.632\pm0.030$&$0.790\pm0.091$\\ \hline
CRO-SVM&$\bf{0.922\pm0.020}$&$\bf{0.827\pm0.020}$&$0.635\pm0.029$&$\bf{0.812\pm0.090}$\\ \hline
CRO-SVM-RL&$0.917\pm0.026$&$\bf{0.828\pm0.022}$&$0.632\pm0.029$&$\bf{0.820\pm0.086}$\\ \hline
CAD-SVM&$0.921\pm0.020$&$\bf{0.828\pm0.019}$&$\bf{0.639\pm0.028}$&$\bf{0.818\pm0.089}$\\ \hline

\end{tabular}
\label{tab_res}
\end{table}

\subsection{Discussions}

Through all the experiments, the CRO-SVM-RL gave as good performance as the CAD-SVM.
As detailed in Appendix~\ref{sec:proof-cad-svm-rl}, we can show the following relations between the CRO-SVM-RL and the CAD-SVM:
\begin{enumerate}
\item{The 0-1-$c$ loss function for the randomly labeled (RL) dataset, in which we randomly relabeled ambiguous samples as positive or negative, reduces to the 0-1-$c$-$d$ loss with $d=\frac{1}{2}-c$.}
\item{For the RL dataset, the MH loss can be regarded as a surrogate loss of the 0-1-$c$-$d$ loss with $d=\frac{1}{2}-c$, and it is calibrated under the conditions of Eq.~(\ref{theo1eq1}).}
\end{enumerate}
Thus, though it is a simple heuristic, the CRO-SVM-RL is essentially equivalent to the CAD-SVM except for that the hyperparameter $d$ is fixed to $\frac{1}{2}-c$. In practice, the CRO-SVM-RL is easier to implement and thus it may be used as an alternative to the CAD-SVM. However, we note that the CRO-SVM-RL alone does not provide rich theoretical insights that we have shown in Section~\ref{sec:cad}.

Our goal for each experiment was minimizing the expected 0-1 loss in the test phase. Therefore, the SVM was naively an appropriate method since the hinge loss was calibrated to the 0-1 loss.
Nevertheless, though the CRO-SVM-RL and the CAD-SVM minimized a surrogate of the 0-1-$c$-$d$ loss in the training phase, they were able to achieve the better accuracy than the SVM in the test phase.
It is suggested that the providing a reject option and incorporating ambiguous training samples could work as a kind of regularization, but further studies will be needed to clarify its mathematical properties.

We note that ambiguous samples are intrinsically hard-to-label samples even by experts, so they usually contain little information for classifying positive and negative samples, and thus we cannot expect large improvement to the binary classification accuracy. Nevertheless, our proposed method achieved statistically significant improvements for some cases.

\section{Conclusion}
In this study, we aimed to reduce the labeling cost and improve the classification accuracy by allowing labelers to give ``ambiguous'' labels for difficult samples. We extended 
a method of classification with reject option
and proposed a novel classification method named the CAD-SVM that uses the 0-1-$c$-$d$ loss.
We derived a surrogate loss for the 0-1-$c$-$d$ loss,
which allowed us to convert the optimization problem into a convex quadratic program.
We carried out numerical experiments and showed that ambiguous labels can be effectively used to improve the classification accuracy.
We also showed that the CRO-SVM-RL, in which we randomly relabeled ambiguous samples to be positive or negative and applied classification with reject option, can be a practical alternative to the proposed method since it is essentially equivalent to the proposed method.

Though our proposed method was based on the SVM, it would be more useful if it can be applied to other models especially deep neural networks.
However, further experimental studies will be needed to confirm if a naive application of the proposed MHA loss works well in practice.
Indeed, it is known that changing models can cause other problems such as overfitting \citep{kiryo2017positive}.
Moreover, for deep neural networks, though we usually use the softmax cross entropy as the loss function, even the 0-1-$c$ loss function has not been extended to the softmax cross entropy.
So, analyzing the influence of changing loss functions is also an important issue to be further investigated.

%Our future study will conduct theoretical analysis of the proposed method such as statistical consistency and the rate of convergence and experimental analysis with more complex models such as neural networks.
In addition to the experimental analysis with more complex models, our future study will conduct theoretical analysis of the proposed method such as statistical consistency and the rate of convergence.
Extending the proposed loss function to semi-supervised problems, imperfect labeling problems or multi-class problems is also a promising direction to be pursued.
%  Other binary classification methods (e.g. logistic regression, neural networks) may be extended to incorporate the ambiguous label. Furthermore, extending the ambiguous label to the other classification problems such as multi-class classification, we can deal with much more real-world problems.
%本研究では、所定のラベルに加えて「分からない」というラベルを許すことにより、ラベリングのコストと分類精度を向上することを目指した。特に、二値分類に着目し、リジェクト付き分類を拡張して、新しいロス関数(0-1-c-d loss)とその代理損失を提案した。更に、代理損失について、ベイズ分類器の符号が0-1-c-d損失と一致するハイパーパラメータを導出した。数値実験を行い、適切なハイパーパラメータのもとで従来手法よりも高精度であることを示した。
%
%今後、提案した損失関数について、収束性などの理論を研究する余地がある。また、SVM以外の二値分類アルゴリズムを基に、類似の損失関数を考えることもできるだろう。更に、マルチクラス分類など他の分類問題に対しても、不明ラベルを考慮した方法が展開できると考えている。

%In this paper, we considered  annotate data that is difficult to discriminate between positive and negative into “ambiguous label” and propose a binary classification algorithm using those data. The proposed algorithm is formulated as an extension of the classification with rejection, and the optimal hyperparameters consistent with the Bayes classifier is derived. Furthermore, numerical experiments are performed using public dataset and in-house dataset and shows that classification accuracy is higher than conventional methods, such as support vector machines and classification with rejection.

\begin{acknowledgements}
%If you'd like to thank anyone, place your comments here
%and remove the percent signs.
The authors would like to thank Yasujiro Kiyota and Momotaro Ishikawa for providing the in-house dataset.
\end{acknowledgements}

% Authors must disclose all relationships or interests that 
% could have direct or potential influence or impart bias on 
% the work: 
%
% \section*{Conflict of interest}
%
% The authors declare that they have no conflict of interest.

% BibTeX users please use one of
\bibliographystyle{spbasic}      % basic style, author-year citations
\bibliography{ambiguous_en_revised}   % name your BibTeX data base

% Non-BibTeX users please use
%\begin{thebibliography}{}
%
% and use \bibitem to create references. Consult the Instructions
% for authors for reference list style.
%
%\bibitem{RefJ}
% Format for Journal Reference
%Author, Article title, Journal, Volume, page numbers (year)
% Format for books
%\bibitem{RefB}
%Author, Book title, page numbers. Publisher, place (year)
% etc
%\end{thebibliography}

\appendix
\section{Proof of~Lemma \ref{lem1}}\label{sec:proof-lemma}
We calculate the expectation value of $L_\mathrm{01cd}$ as follows:
\begin{eqnarray}
&&\mathbb{E}_{y\sim p_0(y|x)}[L_\mathrm{01cd}(h,r,x,y)] \nonumber \\
&=&\pi_+ L_\mathrm{01cd}(h,r,x,+1) + \pi_0 L_\mathrm{01cd}(h,r,x,0) \nonumber + \pi_- L_\mathrm{01cd}(h,r,x,-1) \nonumber \\
&=&\left\{
\begin{array}{ll}
d\pi_0 + \pi_- & \mathrm{if}~\sign(h)=1, \sign(r)=1,\\
d\pi_0 + \pi_+ & \mathrm{if}~\sign(h)=-1, \sign(r)=1, \\
c(\pi_+ + \pi_-) & \mathrm{otherwise}.
\end{array}
\right. \label{lem1eq1}
\end{eqnarray}
Then, the minimum of the expectation value is
\begin{eqnarray}
&&\min_{(h,r)}\mathbb{E}_{y\sim p_0(y|x)}[L_\mathrm{01cd}(h,r,x,y)] \nonumber \\
&=& \min\left(
d\pi_0 + \pi_- , d\pi_0 + \pi_+, c(\pi_+ + \pi_-) \right) \nonumber \\
&=& 
\left\{
\begin{array}{ll}
d\pi_0+\pi_- & \mathrm{if}~\pi_+\geq\frac{ d+(1-c-d)\pi_-}{ c+d},\\
d\pi_0+\pi_+ & \mathrm{if}~\pi_-\geq\frac{ d+(1-c-d)\pi_+}{ c+d},\\
c(\pi_+ + \pi_-)& \mathrm{otherwise}.
\end{array}
\right. \label{lem1eq2}
\end{eqnarray}
From the comparison of Eqs.~(\ref{lem1eq1}) and (\ref{lem1eq2}), the optimal $(h,r)$ subject to $(\pi_+, \pi_-)$ are determined.
\qed

\section{Proof of Theorem~\ref{theo1}}\label{sec:proof-theorem}
\begin{figure}
\centering
\includegraphics[width=7cm,bb=0 150 600 550]{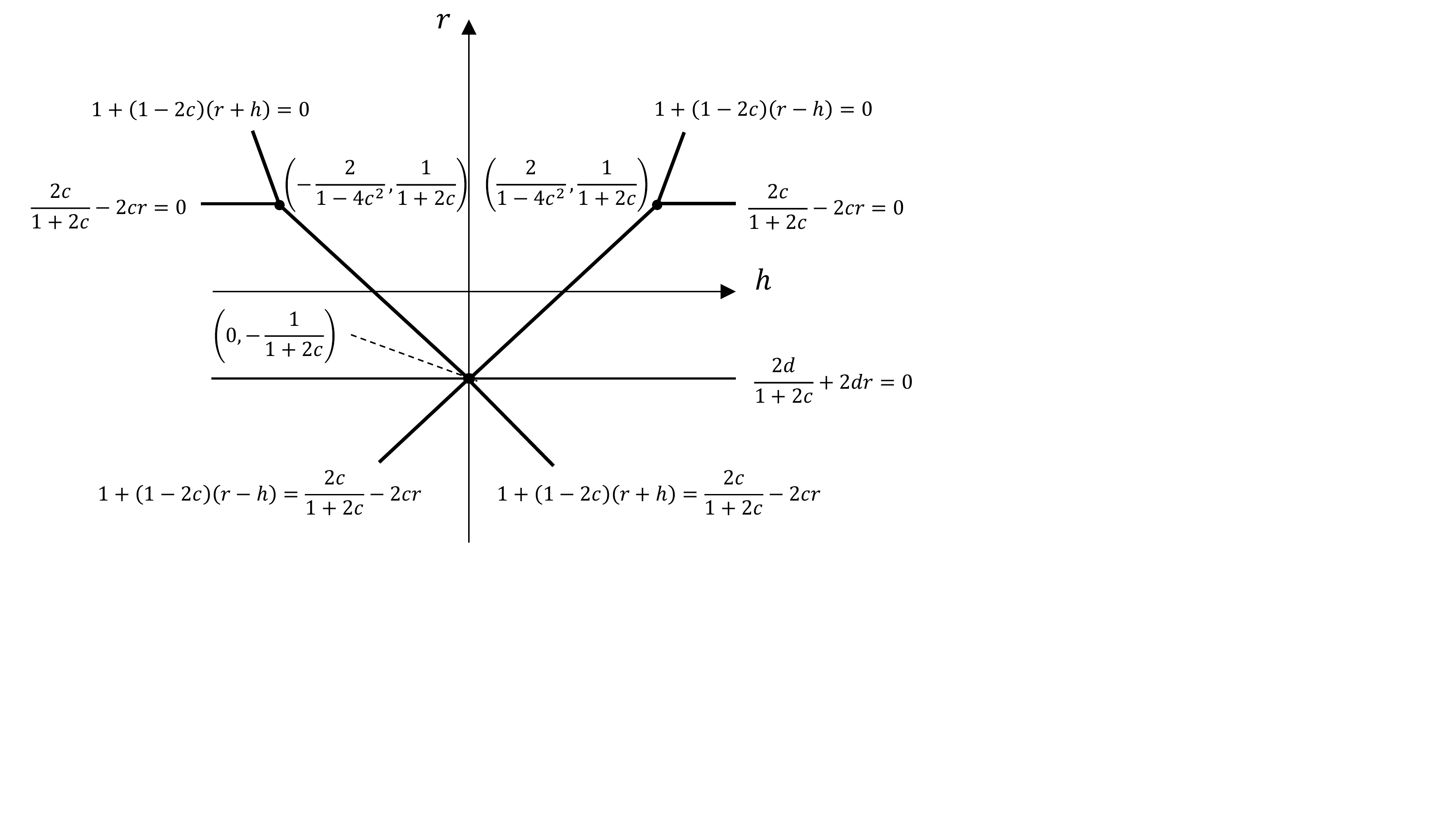}
\caption{The boundary conditions in the linear programming problem of Eq.~(\ref{theo1eq2}).}
\label{theo1fig1}
\end{figure}

At the condition of Eq. (\ref{theo1eq1}),
%\begin{eqnarray}
%L_\mathrm{MHA}(h, r, x, y)&=&y^2 \max \left( 1 + (1-2c)\left(r(x) - yh(x)\right), \frac{2c}{1+2c} - 2c r(x), 0 \right) \nonumber \\*
%&& + (1-y^2) \max \left(\frac{2d}{1+2c} + 2d r(x), 0\right).
%\end{eqnarray}
%Therefore,
the expectation value of the loss is
\begin{eqnarray}
&&\mathbb{E}_{y\sim p_0(y|x)}[L_\mathrm{MHA}(h,r,x,y)] \nonumber \\
&=&\pi_+ \max \left( 1 + (1-2c)\left(r(x) - h(x)\right), \frac{2c}{1+2c} - 2c r(x), 0 \right) \nonumber\\*
&&+\pi_- \max \left( 1 + (1-2c)\left(r(x) + h(x)\right), \frac{2c}{1+2c} - 2c r(x), 0 \right) \nonumber\\*
&&+\pi_0 \max \left(\frac{2d}{1+2c} + 2d r(x), 0\right) . \label{theo1eq2}
\end{eqnarray}
To find the minimum of the expectation value, we derive a linear programming problem considering $r(x), h(x)$ as independent variables.
%\begin{eqnarray}
%&&\min_{(r, h)}\mathbb{E}_{y\sim p_0(y|x)}[L_\mathrm{MHA}(h,r,x,y)] \nonumber \\
%&=& \min_{(r, h, \xi_1, \xi_2, \xi_3)} (\pi_+ \xi_1 + \pi_- \xi_2 + \pi_0 \xi_3) \nonumber \\*
%&&s.t. \left(
%\begin{array}{lll}
%\xi_1 \geq 1+(1-2c)(r-h), & \xi_1 \geq 2c\left( \frac{\disp 1}{\disp 1+2c} - r \right), & \xi_1 \geq 0, \\
%\xi_2 \geq 1+(1-2c)(r+h), & \xi_2 \geq 2c\left( \frac{\disp 1}{\disp 1+2c} - r \right), & \xi_2 \geq 0, \\
%&\xi_3 \geq 2d\left( \frac{\disp 1}{\disp 1+2c} + r \right), & \xi_3 \geq 0
%\end{array}
%\right) \label{theo1lp}
%\end{eqnarray}
As shown in Figure~\ref{theo1fig1}, we can calculate the boundary conditions subject to $(h, r)$ as,
\begin{equation}
\mathbb{E}_{y\sim p_0(y|x)}[L_\mathrm{MHA}(h,r,x,y)] = \left\{ 
\begin{array}{ll}
\frac{4}{1+2c}(d\pi_0+\pi_-) & \mathrm{if}~(h,r)=\left(\frac{ 2}{ 1-4c^2}, \frac{ 1}{ 1+2c}\right), \\
\frac{4}{1+2c}(d\pi_0+\pi_+) & \mathrm{if}~(h,r)=\left(-\frac{ 2}{ 1-4c^2}, \frac{ 1}{ 1+2c}\right), \\
\frac{4}{1+2c}c(\pi_++\pi_-) & \mathrm{if}~(h,r)=\left(0, -\frac{ 1}{ 1+2c}\right).
\end{array}
\right.\label{theo1eq3}
\end{equation}

Thus, we determine the minimizers of $(h, r)$ subject to $(\pi_+, \pi_-)$.
% Since the three values are the multiples of Eq.~(\ref{lem1eq2}), the same condition can be applied.
%\begin{equation}
%\min_{(h,r)}\mathbb{E}_{y\sim p_0(y|x)}[L_\mathrm{MHA}(h,r,x,y)] = \frac{4}{1+2c} \times
%\left\{
%\begin{array}{ll}
%d\pi_0+\pi_- & \left(\pi_+\geq\frac{ d+(1-c-d)\pi_-}{c+d}\right)\\
%d\pi_0+\pi_+ & \left(\pi_-\geq\frac{d+(1-c-d)\pi_+}{c+d}\right)\\
%c(\pi_+ + \pi_-)& (\mathrm{otherwise})
%\end{array}
%\right. \label{theo1eq4}
%\end{equation}
%From the comparison of Eqs.~(\ref{theo1eq3}) and (\ref{theo1eq4}), we can determine the optimal classifiers subject to $(\pi_+, \pi_-)$.
%
\begin{eqnarray}
&(h^*_\mathrm{MHA}, r^*_\mathrm{MHA}) = \argmin_{(h,r)} \mathbb{E}_{y\sim\mathrm{Pr_0}(y|x)}[L_\mathrm{MHA}(h,r,x,y)],&\\
&
\left\{
\begin{array}{ll}
h^*_\mathrm{MHA}=\frac{2}{1-4c^2}>0,\quad r^*_\mathrm{MHA}=\frac{ 1}{ 1+2c}>0 & \mathrm{if}~\pi_+\geq\frac{ d+(1-c-d)\pi_-}{ c+d},\\
h^*_\mathrm{MHA}=-\frac{2}{1-4c^2}<0,\quad r^*_\mathrm{MHA}=\frac{ 1}{ 1+2c}>0 & \mathrm{if}~\pi_-\geq\frac{ d+(1-c-d)\pi_+}{ c+d},\\
h^*_\mathrm{MHA}=0,\quad r^*_\mathrm{MHA}=-\frac{ 1}{ 1+2c}<0 & \mathrm{otherwise}.
\end{array}
\right.&
\end{eqnarray}
\qed

\section{Relation between the CRO-SVM-RL and the CAD-SVM}\label{sec:proof-cad-svm-rl}
For a dataset which contains positive, negative and ambiguous samples, we define a RL (randomly labeled) dataset as a dataset in which ambiguous samples are randomly relabeled into the positive or negative classes.
\begin{theorem} \label{theo2}
The risk of the 0-1-$c$ loss function for the RL dataset is equal to the risk of the 0-1-$c$-$d$ loss with $d=\frac{1}{2}-c$ for the original dataset.
\end{theorem}

{\it Proof}

Let a relabeled label be $z\in\{-1, 1\}$ which satisfies
\begin{equation}
\mathrm{Pr}(z=1|x)=\pi_+(x)+{\scriptstyle\frac{1}{2}}\pi_0(x), \qquad \mathrm{Pr}(z=-1|x)=\pi_-(x)+{\scriptstyle\frac{1}{2}}\pi_0(x).
\end{equation}
Thus, we can calculate the risk of the 0-1-$c$ loss for the relabeled label $z$ as
\begin{eqnarray}
&&\mathbb{E}_{z\sim \mathrm{Pr}(z|x)}[L_\mathrm{01c}(h,r,x,z)] \nonumber \\
&=&\left( \pi_+(x) + {\scriptstyle\frac{1}{2}}\pi_0(x) \right) (1_{h<0}1_{r\geq0} + c1_{r<0}) + \left( \pi_-(x) + {\scriptstyle\frac{1}{2}}\pi_0(x) \right) (1_{h>0}1_{r\geq0} + c1_{r<0}) \nonumber \\
%&=&\pi_+(x) (1_{h<0}1_{r\geq0} + c1_{r<0}) + \pi_-(x)(1_{h>0}1_{r\geq0} + c1_{r<0}) + \pi_0(x) \left(\frac{1}{2}1_{r\geq0} + c 1_{r<0}\right) \nonumber \\
&=&\pi_+(x) (1_{h<0}1_{r\geq0} + c1_{r<0}) + \pi_-(x)(1_{h>0}1_{r\geq0} + c1_{r<0}) + \pi_0(x) \left({\scriptstyle\frac{1}{2}}-c\right) 1_{r\geq0} + c \pi_0(x) \nonumber \\
&=&\mathbb{E}_{y\sim p_0(y|x)}\left[\left.L_\mathrm{01cd}\left(h,r,x,y\right)\right|_{d={\scriptstyle\frac{1}{2}}-c}\right] + \pi_0(x)c.
\end{eqnarray}
Therefore, the risk is equal to that of the 0-1-$c$-$d$ loss for the original label $y$.
Note that the second term in the right-hand side is constant with respect to $h$ and $r$.
\qed

%Theorem~\ref{theo2} indicates that the 0-1-$c$ loss of the RL dataset is equivalent to the 0-1-$c$-$d$ loss of the original dataset.

\begin{theorem} \label{theo3}
The MH loss for the RL dataset is convex and an upper bound of the 0-1-$c$-$d$ loss with $d=\frac{1}{2}-c$ for the original dataset.
\end{theorem}

{\it Proof}

For the ambiguous label, we can calculate the expectation value of the MH loss function over the relabeling process as
\begin{eqnarray}
&&\mathbb{E}_{z\sim \mathrm{Pr}(z|x, y=0)}[L_\mathrm{MH}(h,r,x,z)] \nonumber \\
&=&{\scriptstyle\frac{1}{2}}L_\mathrm{MH}(h,r,x,z=1)+{\scriptstyle\frac{1}{2}}L_\mathrm{MH}(h,r,x,z=-1) \nonumber \\
&=&{\scriptstyle\frac{1}{2}}\max\left(1+{\scriptstyle\frac{\alpha}{2}}\left(r(x)-h(x)\right),c\left(1-\beta r(x)\right),0\right) \nonumber \\
&&+{\scriptstyle\frac{1}{2}}\max\left(1+{\scriptstyle\frac{\alpha}{2}}\left(r(x)+h(x)\right),c\left(1-\beta r(x)\right),0\right) \nonumber \\
&\geq&L_\mathrm{01cd}(h, r, x, y=0).
\end{eqnarray}
Since $L_\mathrm{MH}(h,r,x,z=\pm1)$ is convex, the expectation is also convex.
For positive and negative labels, it can be calculated in the same manner.
\qed

\begin{theorem} \label{theo4}
The expectation of the risk of the MH loss for the RL dataset is calibrated to the 0-1-$c$-$d$ loss with $d=\frac{1}{2}-c$ for the original dataset under the conditions of Eq.~(\ref{theo1eq1}).
\end{theorem}

{\it Proof}

We minimize the expectation of the MH loss with respect to $(r, h)$ for the RL dataset as
\begin{eqnarray}
&&\min_{(h,r)}\mathbb{E}_{z\sim \mathrm{Pr}(z|x)}[L_\mathrm{MH}(h,r,x,z)] \nonumber \\
&=&\min_{(h,r)} \left[ \pi_+(x) L_\mathrm{MH}(h,r,x,z=1) +\pi_-(x) L_\mathrm{MH}(h,r,x,z=-1) \vphantom{\scriptstyle\frac{1}{2}}\right. \nonumber \\
&&\hphantom{\min} \left. + \pi_0(x) {\scriptstyle\frac{1}{2}}\left(L_\mathrm{MH}(h,r,x,z=1)+L_\mathrm{MH}(h,r,x,z=-1)\right) \right]\nonumber \\
&=&\min_{(h,r)}\left[\frac{2}{1+2c}\times \left\{
\begin{array}{ll}
2\pi_+ + \pi_0 & (h=-\frac{2}{1-4c^2}, r=\frac{1}{1+2c}) \\
2\pi_- + \pi_0 & (h=\frac{2}{1-4c^2}, r=\frac{1}{1+2c}) \\
2c & (h=0, r=-\frac{1}{1+2c}) 
\end{array}
\right.\right].
\end{eqnarray}
Then, we calculate the minimizers as
\begin{eqnarray}
&&\argmin_{(h,r)}\mathbb{E}_{z\sim \mathrm{Pr}(z|x)}[L_\mathrm{MH}(h,r,x,z)] \nonumber \\
&=&\left\{
\begin{array}{ll}
(-\frac{2}{1-4c^2}, \frac{1}{1+2c}) & \mathrm{if}~\pi_- \geq (1-2c) + \pi_+, \\
(\frac{2}{1-4c^2}, \frac{1}{1+2c}) & \mathrm{if}~\pi_+ \geq (1-2c) + \pi_-, \\
(0, -\frac{1}{1+2c}) & \mathrm{otherwise}.
\end{array}
\right.
\end{eqnarray}
The derived minimizers are consistent with Lemma~\ref{lem1} with $d=\frac{1}{2} - c $.
\qed
\end{document}